\documentclass[sigconf,nonacm]{acmart}
\AtBeginDocument{%
  }

\usepackage{algorithm}
\usepackage{algpseudocode}
\usepackage{cuted}
\usepackage{tabularx}
\usepackage{float}
\newcommand{\cmark}{Yes}
\setcopyright{none}
\acmDOI{}
\acmISBN{}

\newcommand{\AtikPreprintPageStyle}{%
  \fancyhf{}%
  \fancyhead[R]{\small\itshape Preprint}%
  \fancyfoot[C]{\footnotesize\thepage}%
  \renewcommand{\headrulewidth}{0pt}%
  \renewcommand{\footrulewidth}{0pt}}
\AtBeginDocument{%
  \fancypagestyle{firstpagestyle}{\AtikPreprintPageStyle}%
  \fancypagestyle{standardpagestyle}{\AtikPreprintPageStyle}}

\makeatletter
\g@addto@macro\@mkauthors{%
  \global\setbox\mktitle@bx=\vbox{%
    \unvbox\mktitle@bx
    \vskip 12pt
    \centering
    {\large\bfseries\href{https://github.com/AhmedZeer/atik}{\textcolor{blue}{Project Website.}}\par}
    \vskip 12pt}}
\makeatother

\begin{document}

\title{Atik: A Shared-Mesh Transformer Accelerator for RISC-V}

\author{Ahmed Zeer}
\email{ahmed.zeer@std.yildiz.edu.tr}
\affiliation{%
  \department{Department of Computer Engineering}
  \institution{Yildiz Technical University}
  \city{Istanbul}
  \country{Turkiye}}
\author{Erkan Uslu}
\email{euslu@yildiz.edu.tr}
\affiliation{%
  \department{Department of Computer Engineering}
  \institution{Yildiz Technical University}
  \city{Istanbul}
  \country{Turkiye}}
\renewcommand{\shortauthors}{Zeer and Uslu}
\newcommand{\AtikPDFAuthors}{Ahmed Zeer and Erkan Uslu}

\begin{abstract}
Transformer inference repeatedly invokes dense matrix multiplication and
attention, including its data-dependent softmax normalization. We present
Atik, an open-source RoCC accelerator for RocketChip that executes matrix
multiplication and dense attention through one shared fixed-point MAC mesh.
Atik accepts and returns BF16 tensors, uses a streaming online-softmax schedule
that never materializes the full score matrix, and is invoked through a small
descriptor-based ABI. We evaluate three meshes cycle-by-cycle on FireSim and
report standalone Sky130 physical-design area. Across 137 logged FireSim test
records, all tests pass their benchmark tolerance with zero reported output
mismatches. Relative to the same scalar Rocket BF16 reference, the 8$\times$8
configuration reaches 31.5$\times$ peak matmul speedup and 225.8$\times$ peak
attention speedup. Hybrid projections for the single-layer workload-derived C
models, using directly measured accelerator-kernel cycles, give aggregate
speedups of 44.2$\times$, 77.1$\times$, and 32.3$\times$ for ViT, TinyBERT,
and GPT-2 prefill, respectively; these are not full-model hardware replays.
\end{abstract}



\maketitle
\pagestyle{standardpagestyle}
\hypersetup{pdfauthor={\AtikPDFAuthors}}

\section{Introduction}

\begin{figure}[t]
  \centering
  \includegraphics[width=\linewidth]{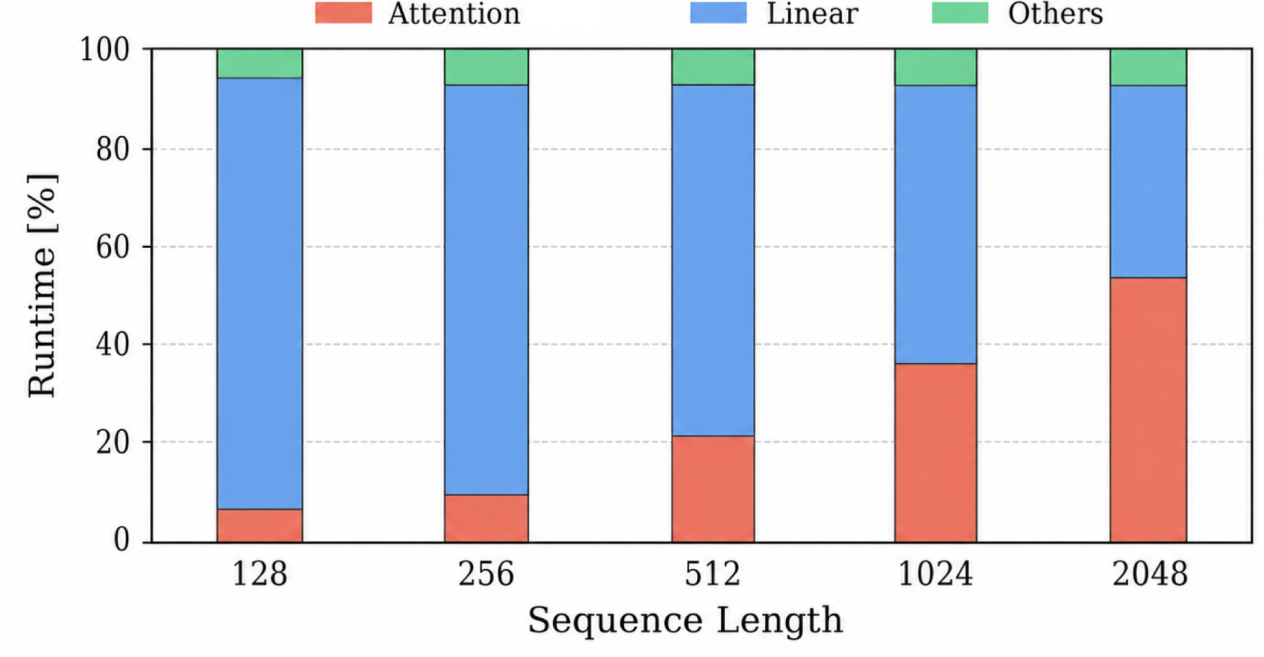}
  \caption{Runtime distribution of operations in the evaluated transformer
    decoder workload.}
  \Description{Bar chart showing the relative runtime contribution of major
    transformer operations.}
  \label{fig:trans}
\end{figure}

The rapid, cross-domain adoption of transformer models has made a small number
of compute kernels disproportionately important. Nearly every modern language
and vision model is built on the same backbone, so the same operations dominate
inference cost across otherwise unrelated workloads. Two kernels stand out:
dense matrix multiplication, which drives the linear projections and
feed-forward layers, and the attention mechanism, whose softmax normalization
introduces data-dependent, non-linear work that general-purpose datapaths
handle poorly. Where the bottleneck lies is not fixed. On a general-purpose
core without specialized vector or matrix units, both dense matrix operations
and software softmax can be costly. Dedicated matrix hardware can reduce the
relative cost of matrix multiplication, exposing normalization and attention
data movement as additional bottlenecks
\cite{onlinesoftmax,softermax,vexp,ita}. Their relative costs depend on the
platform, model, and sequence length. An accelerator that addresses only one
of the two kernels leaves a substantial fraction of the workload on the slow
path. Figure~\ref{fig:trans} summarizes this split for the transformer decoder
workload used in our study.

\begin{figure}[t]
  \centering
  \includegraphics[width=\linewidth]{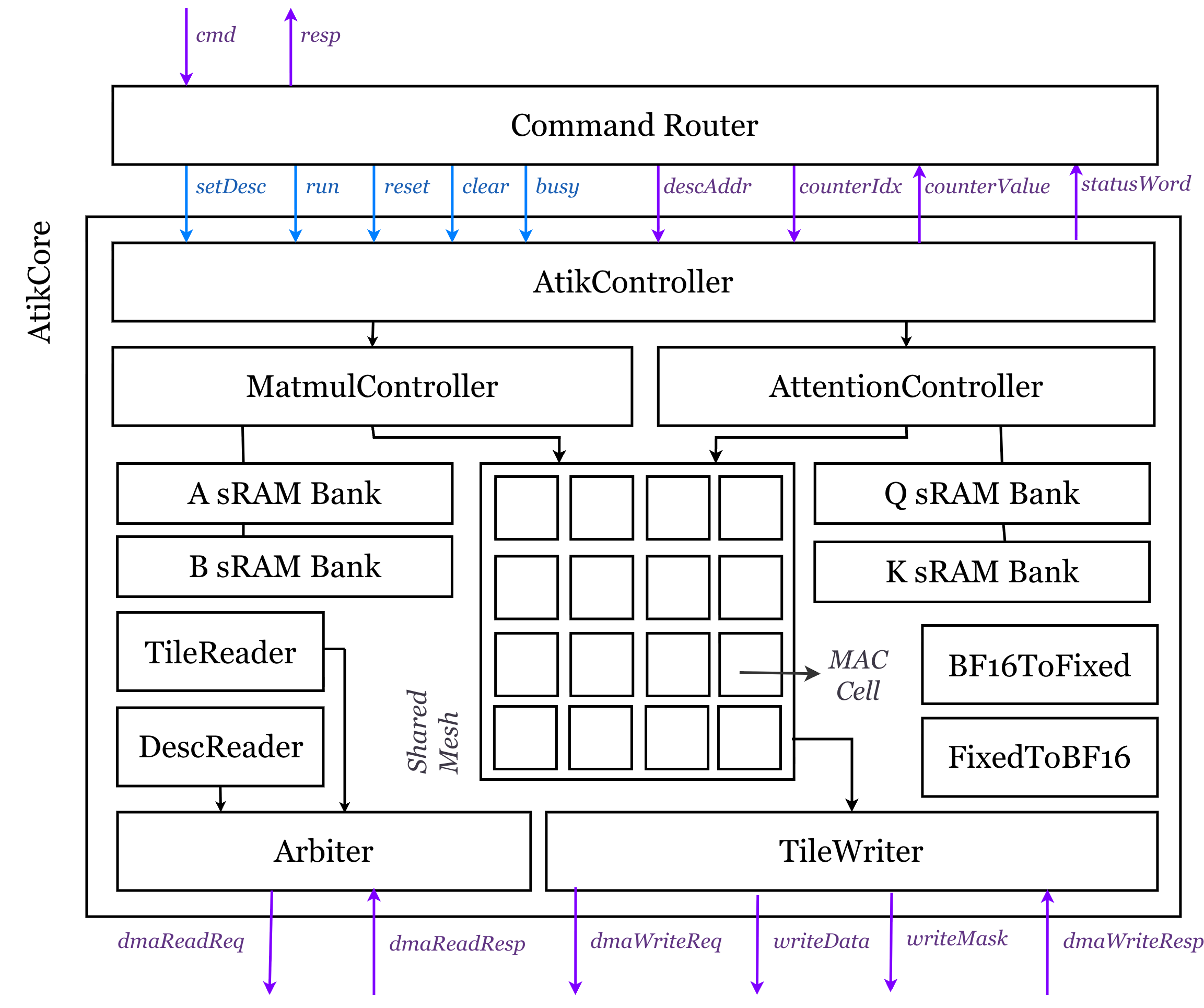}
  \caption{AtikCore block diagram. A RoCC command router drives a top-level
    controller that dispatches to the matmul or attention controller. Both
    share a single MAC mesh, the BF16$\leftrightarrow$fixed-point converters,
    the SRAM tile banks, and the tile-DMA read/write path.}
  \Description{Block diagram of the Atik accelerator core showing a command
    router feeding a controller, which dispatches to matmul and attention
    controllers that share a MAC mesh, SRAM banks, numeric converters, and a
    tile DMA arbiter and writer.}
  \label{fig:archi}
\end{figure}

The architecture community has responded with a large body of specialized
hardware, but the resulting designs tend to cluster into a few shapes, each
with a practical limitation. Many are standalone ASICs that accelerate
attention or softmax as standalone blocks, rather than demonstrating the
RoCC integration and shared memory-system interface targeted here
\cite{a3,elsa,sanger,energon}. Other work uses architectural full-system
simulation rather than an FPGA execution of the integrated RTL
\cite{ticsat}. A third group targets a single narrow
operation, such as adding an exponential unit to a floating-point pipeline
\cite{vexp}. Finally, most of these designs commit to quantized or integer
datatypes such as INT8 \cite{ita,swiftron}, or approximate softmax with
lookup tables or powers of two, introducing numerical trade-offs that must be
evaluated at the workload level \cite{softermax,sole}. The mature open-source
DNN accelerators that predate the
transformer era, such as Gemmini \cite{gemmini}, were optimized for
convolutional and matrix workloads rather than a dedicated streaming attention
pipeline. Atik instead exposes a BF16 tensor interface around a fixed-point
datapath for both matrix multiplication and attention.

\subsection{Positioning}
Table~\ref{tab:positioning} compares the integration, arithmetic, and evaluation
scope of Atik with representative accelerators. Atik combines a RoCC interface,
a shared mesh for matrix multiplication and attention, and integrated RTL
evaluation on FireSim.

\begin{strip}
\begin{minipage}{\textwidth}
  \captionof{table}{Positioning of Atik against representative transformer-related
    accelerators. \emph{Coupling} distinguishes a standalone ASIC, an
    instruction-set extension (ISA-ext), and a tightly-coupled accelerator
    attached to a general-purpose core (TCA). \emph{Evaluation method} names
    the platform and evidence type used by each paper; it is not a quality
    ranking. ``Yes'' indicates that the paper reports the corresponding
    capability or releases the design; a dash indicates that it does not.}
  \label{tab:positioning}
  \small
  \setlength{\tabcolsep}{4pt}
  \renewcommand{\arraystretch}{1.2}
  \begin{tabularx}{\textwidth}{@{}lcl>{\raggedright\arraybackslash}X>{\raggedright\arraybackslash}p{1.65cm}c>{\raggedright\arraybackslash}p{1.4cm}c@{}}
    \toprule
    Work & Open & Coupling & Evaluation method & Datatype & MatMul & \shortstack[l]{Softmax/\\attention}
      & \shortstack{No\\retrain} \\
    \midrule
    Gemmini \cite{gemmini}        & \cmark & TCA (RoCC) & FPGA performance; ASIC synthesis
      & INT/FP & \cmark & --- & \cmark \\
    TiC-SAT \cite{ticsat}         & \cmark & TCA        & gem5-X full-system simulation
      & INT & \cmark & \cmark & \cmark \\
    VEXP \cite{vexp}              & --- & ISA-ext    & ModelSim; GF12 synthesis/P\&R & BF16
      & --- & softmax (exp) & \cmark \\
    ITA \cite{ita}                & \cmark & ASIC       & GF22FDX synthesis/P\&R; gate-level power & INT8
      & \cmark & \cmark & --- \\
    Softermax \cite{softermax}    & --- & ASIC       & HLS area/energy; PyTorch accuracy & FP (approx.)
      & --- & softmax & --- \\
    A$^3$ \cite{a3}               & --- & ASIC       & 40nm test chip; CPU/GPU workloads & FP
      & --- & attention & \cmark \\
    ELSA \cite{elsa}              & --- & ASIC       & 40nm synthesis/area/power; CPU/GPU workloads & FP
      & --- & attention & \cmark \\
    Sanger \cite{sanger}          & \cmark & ASIC       & Verilator; DC area/power; CPU/GPU workloads & INT
      & --- & attention & --- \\
    Energon \cite{energon}        & --- & ASIC       & Verilator; DC area/power; DRAMPower & mixed
      & --- & attention & \cmark \\
    \textbf{Atik (this work)}     & \cmark & TCA (RoCC) & FireSim cycles; Sky130 area estimate
      & \mbox{BF16 I/O;}\newline fixed & \cmark & \cmark & --- \\
    \bottomrule
  \end{tabularx}
\end{minipage}
\end{strip}

For Atik, the datatype entry describes the external interface and internal
arithmetic separately. Benchmark tolerance checks do not establish task-level
accuracy without retraining, so the corresponding entry is left unasserted.

The RISC-V ecosystem offers an underused integration point for exactly this
kind of accelerator. RocketChip \cite{rocketSoC} exposes the Rocket Custom
Coprocessor (RoCC)
interface, through which a tightly-coupled accelerator can be attached
alongside the main core and access the system memory hierarchy. Because
RocketChip is generated with Chisel \cite{chisellang} inside the
Chipyard framework \cite{chipyard}, the same design can be simulated at RTL,
run cycle-accurately on an FPGA through FireSim \cite{firesim}, and pushed
through a physical-design flow \cite{hammer}. This makes RoCC an attractive
middle ground: an accelerator that is neither as heavyweight and isolated as a
from-scratch ASIC, nor as limited as a pure instruction-set extension, and that
can be exercised by real software running on a real core.

We present \emph{Atik}, an open-source transformer accelerator that occupies
exactly this middle ground. Atik attaches to RocketChip over RoCC and
accelerates both matrix multiplication and attention with BF16 input/output. Its
central design decision is a \emph{shared} fixed-point MAC mesh that serves all
three heavy operations---matmul ($C \mathrel{+}= A\,B$), attention scoring
($\mathrm{score} \mathrel{+}= Q\,K^\top$), and probability-value accumulation
($O \mathrel{+}= P\,V$)---rather than provisioning separate datapaths. Around
this mesh, attention is executed with a streaming online-softmax schedule that
never materializes the full score matrix, keeping only per-row running maxima
and denominators as live state. Software drives the accelerator through a
deliberately small descriptor-based ABI: it fills a descriptor with tensor
pointers and shapes, issues a set-descriptor and a run command, and polls for
completion, while the hardware owns all tiling, DMA, packing, conversion,
softmax bookkeeping, and writeback. Atik is validated on two fronts that most
prior work treats separately: it runs cycle-accurately on an AWS~F2 FPGA
through FireSim, and it has a preliminary standalone Sky130 area study.

This paper makes the following contributions:
\begin{itemize}
  \item \textbf{A tightly-coupled, software-callable accelerator on a real
    core.} Atik integrates directly into RocketChip over RoCC, is invoked from
    ordinary C software, and shares the core's memory hierarchy, rather than
    existing as an isolated ASIC or a non-cycle-accurate simulation model.
  \item \textbf{A unified design for both transformer bottlenecks.} A single
    shared fixed-point MAC mesh accelerates dense matmul, QK scoring, and the
    accumulation of probabilities against values, while a streaming softmax
    schedule handles the nonlinear normalization without ever storing the full
    score matrix. Atik therefore targets matrix multiplication and attention
    together, rather than either one in isolation.
  \item \textbf{BF16-compatible interface with measured numerical checks.}
    Atik consumes and produces row-major BF16 tensors, while its mesh and
    softmax datapath use fixed point and lookup tables. We report benchmark
    tolerance checks rather than claiming bit-exact BF16 arithmetic.
  \item \textbf{System-level evaluation and preliminary area evidence.} We
    evaluate Atik cycle-by-cycle on AWS~F2 through FireSim and report a
    standalone Sky130 area estimate across 2$\times$2, 4$\times$4, and
    8$\times$8 mesh configurations.
  \item \textbf{An open-source artifact.} The complete design, software, and
    hardware/VLSI flow are released to the community.
\end{itemize}

\section{Background and Related Work}

Systolic arrays are a standard approach to dense matrix multiplication
\cite{sysArrKung,googletpu,gemmini}, while online softmax avoids materializing
full score rows \cite{onlinesoftmax,softermax}. Transformer accelerators span
tightly-coupled systems such as TiC-SAT \cite{ticsat}, narrow exponential units
such as VEXP \cite{vexp}, and standalone attention ASICs
\cite{ita,a3,elsa,sanger,energon}. Table~\ref{tab:positioning} summarizes the
dimensions relevant here. Atik's narrow contribution is not a claim of
bit-exact floating-point execution or superiority to these systems; it is the
combination of a RoCC-attached interface, a shared mesh for matmul and both
attention products, and cycle-level evaluation of the integrated design.

\section{Atik Architecture}

Atik is a RoCC-attached accelerator for two operations: dense matrix
multiplication and attention. Its guiding principle is reuse: rather than
provisioning a separate datapath per operation, the design routes matmul, the
QK score computation, and the probability--value accumulation through a
single shared fixed-point MAC mesh, and moves all data with one shared tile-DMA
engine and one set of SRAM tile buffers. Figure~\ref{fig:archi} shows the
resulting core. Software interacts with Atik only through a small descriptor
and a handful of RoCC commands; the hardware owns tiling, data packing, DMA
scheduling, numeric conversion, softmax bookkeeping, and writeback.

\subsection{RoCC Integration and Control Flow}
Atik is instantiated as a Rocket Custom Coprocessor on the \texttt{custom0}
opcode and exposes a TileLink client port, through which it issues
\texttt{Get} and \texttt{Put} transactions to the RocketChip memory system.
Software drives the
accelerator with a minimal command set: it fills an 80-byte descriptor in
memory, passes the descriptor address with \texttt{set\_desc}, and starts
execution with \texttt{run}; additional commands read the status word and the
performance counters. The descriptor carries the operation code
(matmul, attention, or causal attention), the tensor base addresses and leading
dimensions, the problem dimensions, and, for attention, the BF16 scale factor.
On \texttt{run}, a descriptor reader fetches the descriptor over the shared read
port, the top-level controller decodes the operation, and it hands off to either
the matmul or the attention controller. Software then polls the status word
until the \texttt{busy} bit clears. Crucially, the software-visible interface
never exposes tiling, packing, or internal addresses: the same descriptor
describes an arbitrarily large problem, and the hardware streams it in tiles.

\subsection{Shared MAC Mesh}
\label{sec:sharedmesh}
The compute core is an $R\times C$ mesh of fixed-point multiply--accumulate
cells (Fig.~\ref{fig:archi}), configurable as $2\times2$, $4\times4$, or
$8\times8$. Each cycle the mesh is driven by a row vector of $R$ operands and a
column vector of $C$ operands, and every cell $(r,c)$ computes the outer-product
update $\mathit{acc}_{r,c} \mathrel{+}= a_r \cdot b_c$ against a live
accumulator tile held in registers. The accumulator remains resident across the
entire reduction dimension of a tile, so a large $K$ (or key/value length) is
consumed as a sequence of depth-$K_T$ chunks without ever spilling partial
sums to memory.

The same mesh serves all three heavy computations. For matmul it evaluates
$C \mathrel{+}= A\,B$; for attention it first evaluates the score tile
$\mathrm{score} \mathrel{+}= Q\,K^\top$, and later folds the normalized
probabilities into the output as $O \mathrel{+}= P\,V$. A single multiplexer at
the mesh inputs selects whether the operands come from the matmul or the
attention controller, so the largest arithmetic structure in the design is paid
for once and amortized across every operation. This is the central difference
from designs that instantiate independent matmul and attention engines. The
preliminary area study in Section~\ref{sec:vlsi} measures this amortization as
the mesh grows.

\subsection{Numeric Datapath}
External tensors are row-major BF16. Internally, Atik computes in fixed point: on the way into
the mesh, a \texttt{BF16toFixed} unit converts each element to a 24-bit signed
fixed-point value, products accumulate in a 64-bit signed accumulator, and a
\texttt{FixedToBF16} unit converts results back to BF16 for writeback. The
attention softmax path uses a wider 32-bit unsigned fixed-point format for
exponentials and running sums. Exponentiation and reciprocal are implemented as
small lookup tables---a 256-entry table evaluating $e^{-x}$ over $[0,16]$ and
a 256-entry reciprocal table for final normalization---so the design avoids
iterative dividers. These conversions and lookup tables are approximations;
Section~\ref{sec:setup} reports the implemented tolerance-based checks.

\subsection{Tile DMA and Local Buffering}
A single small tile-DMA engine decouples memory movement from compute. It is
parameterized only by a base address, row and column counts, a stride, and a
direction, which is sufficient to gather a row-major BF16 tile from memory or
scatter one back. The matmul and attention controllers arbitrate for the shared
reader and writer, and the descriptor reader shares the same read port, so the
entire core presents just one read and one write stream to the memory system.
Loaded tiles land in SRAM-backed buffers---per-row banks for the $A$/$Q$ operand
and per-column banks for the $B$/$K$ operand---from which the controllers read
back $K_T$-deep chunks, convert them to fixed point, and drive the mesh. Only
the small live state (the accumulator tile and, for attention, the per-row
maximum and denominator) is kept in registers. This explicit, tile-granular
data movement keeps the memory interface simple and makes the accelerator's
traffic predictable, which the on-chip performance counters expose directly for
analysis.

\section{Attention Dataflow}
\label{sec:attention}

Attention computes $O = \mathrm{softmax}\!\left(QK^\top\!/\sqrt{d_k}\right)V$.
The naive formulation materializes the full $q_{\text{rows}}\times
kv_{\text{rows}}$ score matrix, which is both memory-hungry and a poor match for
a small tightly-coupled accelerator. Atik instead follows the streaming
online-softmax formulation \cite{onlinesoftmax}: it walks the key/value axis in
chunks, and for each chunk it computes a score tile, updates a running per-row
maximum and denominator, and immediately folds the result into a resident output
accumulator. The full score matrix is never stored---only one
$\mathit{MR}\times\mathit{KC}$ score tile and its exponentiated copy exist at any
time. Algorithm~\ref{alg:attention} gives the resulting schedule.

The two matrix products in attention, $QK^\top$ and the probability--value
product, run on the shared MAC mesh, while softmax bookkeeping runs on one
scalar lane with exponential and reciprocal lookup tables.

\begin{algorithm}[t]
\caption{Atik tiled online-softmax attention. $MR$/$NC$ are the mesh row/column
  tile sizes, $KC$ the key/value chunk, and $KT$ the internal reduction depth.
  The score matrix is never fully materialized; $QK^\top$ and the
  probability--value product run on the shared mesh, and the softmax update runs
  on the scalar lane.}
\label{alg:attention}
\begin{algorithmic}[1]
\For{$q_0 = 0$ to $q_{\text{rows}}$ step $MR$}
  \For{$v_0 = 0$ to $\mathit{val\_cols}$ step $NC$}
    \State $\mathit{rowMax}\gets-\infty;\ \ \mathit{rowSum}\gets0;\ \ \mathit{outAcc}\gets0$
    \For{$kv_0 = 0$ to $kv_{\text{rows}}$ step $KC$}
      \State $\mathit{score}\gets0$ \Comment{$QK$ on the shared mesh}
      \For{$d_0 = 0$ to $d_k$ step $KT$}
        \State load $Q,K$ tiles;\ \ $\mathit{score}\mathrel{+}=Q\,K^\top$
      \EndFor
      \State $\mathit{score}\mathrel{*}=\mathit{scale}$; apply causal mask \Comment{scalar lane}
      \State $\mathit{newMax}\gets\max(\mathit{rowMax},\ \textsc{rowmax}(\mathit{score}))$
      \State $\mathit{oldScale}\gets\exp(\mathit{rowMax}-\mathit{newMax})$
      \State $\mathit{outAcc}\mathrel{*}=\mathit{oldScale}$
      \State $\mathit{scoreExp}\gets\exp(\mathit{score}-\mathit{newMax})$
      \State $\mathit{rowSum}\gets\mathit{rowSum}\cdot\mathit{oldScale}+\textsc{sum}(\mathit{scoreExp})$
      \State load $V$ tile;\ \ $\mathit{outAcc}\mathrel{+}=\mathit{scoreExp}\cdot V$ \Comment{$PV$ on the shared mesh}
      \State $\mathit{rowMax}\gets\mathit{newMax}$
    \EndFor
    \State $\mathit{outAcc}\mathrel{*}=\textsc{reciprocal}(\mathit{rowSum})$ \Comment{normalize}
    \State write $\mathrm{bf16}(\mathit{outAcc})$ to $O$ tile
  \EndFor
\EndFor
\end{algorithmic}
\end{algorithm}

For each key/value chunk, the mesh computes $QK^\top$ over depth-$K_T$ slices.
The scalar lane applies scaling and the causal mask, updates the running maximum
and denominator, and rescales the live output. The resulting probability tile is
then multiplied by $V$ on the same mesh. After the final chunk, reciprocal-based
normalization and BF16 conversion precede writeback. The implementation uses one
scalar lane, an explicit area-first choice that serializes this non-linear phase.

The value-column loop encloses the key/value loop in
Algorithm~\ref{alg:attention}. Consequently, the score and softmax computations
are repeated for each output-column tile when $\mathit{val\_cols}>NC$.
This schedule bounds the resident output state at the cost of recomputation
and additional input traffic.

\section{Experimental Setup}
\label{sec:setup}

\subsection{Configurations and Platform}
We evaluate three Atik instances that differ only in mesh size: $2\times2$,
$4\times4$, and $8\times8$, corresponding to 4, 16, and 64 MAC lanes. All three
expose the identical descriptor ABI and run the same unmodified benchmark
binaries; only the generated hardware changes. Each instance is attached over
RoCC to one Rocket core in a Chipyard-generated SoC with a 128-bit system bus. Functional
bring-up and debugging use a Verilator RTL simulator, while all reported
performance numbers come from cycle-accurate FPGA emulation with FireSim on AWS
F2. The logged runs use an effective target frequency of approximately 10 MHz.
Because FireSim is
cycle-accurate, every measurement is expressed in \emph{target clock cycles},
read from the RISC-V cycle CSR and from Atik's on-chip performance counters;
these cycle counts are independent of the host FPGA clock, so speedups are
reported as cycle ratios rather than wall-clock times.

\subsection{Baseline}
The baseline is the same Rocket core running a BF16 software reference
implementation of each kernel, on the same SoC and memory system. This isolates
the effect of the accelerator: the comparison is Rocket-with-Atik against
Rocket alone, not against a different platform. For every accelerated kernel,
the hardware output is compared with the software reference using a
workload-specific absolute-error tolerance. The archived FireSim logs contain
137 passing result records with zero reported mismatches; the largest logged
absolute difference is 0.0625. These checks establish benchmark-level numerical
agreement, not task accuracy or bit-exact equivalence.

\subsection{Workloads}
The benchmark suite spans two standalone kernels and three workload-derived C
transformer models, chosen to exercise both overhead-dominated and
compute-dominated regimes.

\emph{Standalone kernels.} The matmul benchmark multiplies an $M\times K$
matrix by a $K\times N$ matrix over a range of $M$, $N$, and $K$, and the attention
benchmark computes both non-causal and causal attention over query count $Q$,
key/value count $KV$, key/query dimension $D$, and value dimension $V$. These
isolate the accelerator's behavior without surrounding model bookkeeping.

\emph{Workload-derived C models.} TinyBERT, a Vision Transformer (ViT), and the
GPT-2 prefill stage are represented by small C reference models that preserve
the relevant layer structure while varying sequence length, model width, head
count, and final projection size. Multi-head attention is
decomposed into per-head $QK$, softmax, and $PV$ operations; Atik accelerates
these repeated attention and projection kernels while the software harness keeps
the tensor layout and model-level accounting consistent. Numerical fidelity is
tracked by comparing three outputs---the original ground-truth result, the C
reference, and the Atik result---so that the C port is validated against the
model before the accelerator is judged against the C port.

\subsection{Measurement Methodology}
For the standalone kernels, the software reference is timed with the RISC-V
cycle counter. Atik is measured two ways: the same counter surrounds the RoCC
call for end-to-end cycles, while \texttt{TOTAL\_CYCLES} records internal
accelerator cycles. We report algorithmic FLOP/cycle from both denominators;
these counts describe the equivalent matrix work, although Atik executes
fixed-point arithmetic internally. Matmul work is
$2MNK$ FLOPs. For attention we report \emph{attention-matmul FLOPs},
$2QKV(D+V)$, for $QK^\top$ and $PV$; softmax remains in the cycle denominator
but has no arbitrary FLOP-equivalent count. Replaying the full workload-derived
C models on both paths would be expensive in FPGA simulation, so for those workloads we
profile the full CPU execution, separately profile the stages that Atik
replaces, and form a hybrid projected cost as
\begin{equation}
c_{\text{acc}} = c_{\text{cpu,total}} - c_{\text{cpu,replaced}}
               + c_{\text{hw,replacement}},
\label{eq:hybrid}
\end{equation}
where $c_{\text{cpu,replaced}}$ is the measured CPU cost of exactly the kernels
Atik takes over and $c_{\text{hw,replacement}}$ is their measured cost on Atik.
This is a hybrid projection rather than a full-model hardware replay: it uses
directly measured hardware-kernel cycles while assuming the unaccelerated
remainder is unchanged. We report two
summaries per workload and configuration. The \emph{aggregate} speedup sums all
baseline cycles and all accelerated cycles over the matching problem sizes and
takes their ratio, $\sum c_{\text{cpu}} / \sum c_{\text{hw}}$, which weights
cases by their cycle costs; the \emph{peak} speedup reports the
single best observed case, which reflects how well the design amortizes control,
DMA, and setup overhead once the problem is large enough to keep the mesh busy.

\section{Performance Results}
\label{sec:results}

\begin{figure}[t]
  \centering
  \includegraphics[width=\linewidth]{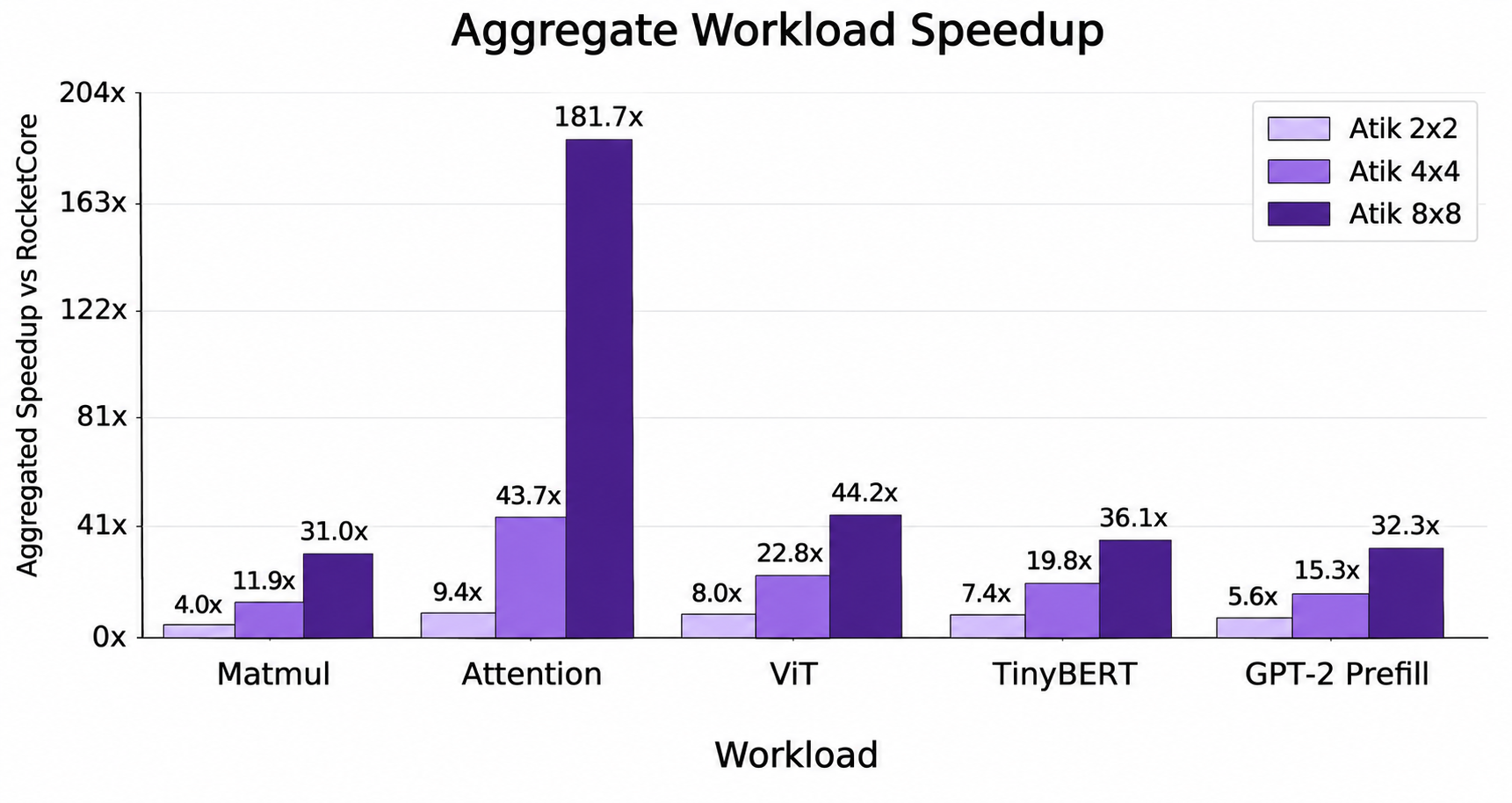}
  \caption{Aggregate speedup, $\sum c_{\text{cpu}}/\sum c_{\text{hw}}$, over the
    Rocket baseline for the $2\times2$, $4\times4$, and $8\times8$
    configurations. Kernel results are directly measured; ViT, TinyBERT, and
    GPT-2 prefill use hybrid projections. The dashed line marks the $1\times$
    Rocket baseline.}
  \Description{Grouped bar chart over Matmul, Attention, ViT, TinyBERT, and GPT-2
    prefill, with one bar per Atik configuration (2x2, 4x4, 8x8) relative to a
    RocketCore baseline of one. Aggregate speedups grow with mesh size and are
    highest for attention.}
  \label{fig:aggregate}
\end{figure}

Figure~\ref{fig:aggregate} summarizes aggregate speedups over the Rocket
baseline. Table~\ref{tab:efficiency} reports representative kernel efficiency,
while Tables~\ref{tab:vit}, \ref{tab:gpt2}, and~\ref{tab:tinybert} report the
hybrid projected workload speedups. Across the matched cases, the larger
meshes deliver the highest speedups.

\subsection{Overall Speedups}
Table~\ref{tab:efficiency} reports how much useful arithmetic is completed per
target cycle, with both end-to-end and internal accelerator cycle counts.

\begin{table}[H]
  \centering
  \caption{Direct 8$\times$8 kernel efficiency from FireSim logs. End-to-end
    cycles include the RoCC call; internal cycles are \texttt{TOTAL\_CYCLES}.
    Attention counts only its two matrix products in the FLOP numerator.}
  \label{tab:efficiency}
  \small
  \begin{tabular}{lrrr}
    \toprule
    Case & CPU & Atik e2e & Atik int. \\
    & FLOP/cyc. & FLOP/cyc. & FLOP/cyc. \\
    \midrule
    GEMM $64^3$ & 0.1151 & 3.6195 & 3.6224 \\
    Attn. $128^4$ & 0.002210 & 0.4100 & 0.4100 \\
    \bottomrule
  \end{tabular}
\end{table}

For the $64\times64\times64$ GEMM, Rocket performs 0.1151
FLOP/cycle, while Atik performs 3.6195 FLOP/cycle including the complete RoCC
call. The ratio is 31.5$\times$, the same speedup reported above. The 8$\times$8
mesh can issue at most 128 FLOP/cycle, so this measured case reaches 2.83\% of
the ideal arithmetic rate after memory movement, conversion, control, and
writeback are included. For $Q{=}KV{=}D{=}V{=}128$ attention, the corresponding
rate is 0.4100 FLOP/cycle for the two matrix products. The softmax is included
in the measured cycles but not converted into FLOPs, because LUT lookups and
scalar reductions have no standard FLOP equivalent. The internal counters also
show where time goes: 16,220,220 DMA-read cycles, 2,502,656 mesh-compute cycles,
and 1,945,600 softmax cycles. These counters are overlapping component
counters, so they are not added together.

On the standalone kernels, the $8\times8$ design reaches $31.0\times$ aggregate
and $31.5\times$ peak on matmul, and $181.7\times$ aggregate with a $225.8\times$
peak on attention. Attention benefits the most because all three of its heavy
components---the $QK$ score computation, the softmax normalization, and the
probability--value product---are offloaded together: on the baseline, the
scalar, data-dependent softmax is especially expensive, so replacing the whole
attention pipeline removes a bottleneck that a matmul-only accelerator would
leave in place. This is the concrete payoff of targeting both transformer
bottlenecks with one design rather than one bottleneck in isolation.

The workload-derived C models show the same ordering with more moderate
magnitudes, because they interleave accelerated kernels with model-level
bookkeeping that stays on the core. For the $8\times8$ configuration, estimated
aggregate speedups are $44.2\times$ for ViT, $77.1\times$ for TinyBERT, and
$32.3\times$ for GPT-2 prefill; their peak cases reach $50.5\times$,
$96.0\times$, and $36.3\times$, respectively.

Table~\ref{tab:vit} shows ViT speedups increasing from $26.6\times$ to
$50.5\times$ on the $8\times8$ mesh across the evaluated cases. The largest
case also benefits most from widening the mesh, rising from $8.4\times$ on
the $2\times2$ configuration to $50.5\times$ on the $8\times8$ configuration.

\begin{table}[H]
  \centering
  \caption{Per-case hybrid projected ViT speedup over the Rocket baseline. $S$ is the post-patch
    token count, $D$ the model dimension, $H$ the head count, and MLP the
    feed-forward hidden size (all cases use 10 output classes).}
  \label{tab:vit}
  \begin{tabular}{rrrrccc}
    \toprule
    $S$ & $D$ & $H$ & MLP & $2\times2$ & $4\times4$ & $8\times8$ \\
    \midrule
    64  & 64  & 4 & 128 & 7.3 & 16.9 & 26.6 \\
    64  & 128 & 4 & 128 & 7.1 & 19.6 & 36.8 \\
    121 & 128 & 4 & 256 & 8.4 & 24.9 & 50.5 \\
    \bottomrule
  \end{tabular}
\end{table}

Table~\ref{tab:gpt2} reports GPT-2 prefill speedups of $14.4\times$ to
$36.3\times$ on the $8\times8$ mesh. The smaller prefill cases amortize setup
and data-movement costs less effectively than the largest evaluated case.

\begin{table}[H]
  \centering
  \caption{Per-case hybrid projected GPT-2 prefill speedup over the Rocket baseline. $S$ is the
    sequence length, $D$ the model dimension, $H$ the head count, and $V$ the
    vocabulary projection size (FFN hidden size 128/256).}
  \label{tab:gpt2}
  \begin{tabular}{rrrrccc}
    \toprule
    $S$ & $D$ & $H$ & $V$ & $2\times2$ & $4\times4$ & $8\times8$ \\
    \midrule
    16 & 64  & 4 & 128 & 4.3 & 8.9  & 14.4 \\
    32 & 128 & 4 & 256 & 5.4 & 14.1 & 28.6 \\
    64 & 128 & 4 & 256 & 5.7 & 16.4 & 36.3 \\
    \bottomrule
  \end{tabular}
\end{table}

Table~\ref{tab:tinybert} reports the TinyBERT cases, including the highest
projected workload speedup of $96.0\times$ for 128 tokens and model dimension
256 on the $8\times8$ mesh. Missing configurations are marked explicitly;
cross-configuration aggregate comparisons use only matched cases.

\begin{table}[H]
  \centering
  \caption{Per-case hybrid projected TinyBERT speedup over the Rocket baseline ($S$ tokens, model
    dimension $D$, $H$ heads; 2 output classes). ``---'' marks a case not run
    for that configuration; aggregate comparisons use only matched cases.}
  \label{tab:tinybert}
  \begin{tabular}{rrrccc}
    \toprule
    $S$ & $D$ & $H$ & $2\times2$ & $4\times4$ & $8\times8$ \\
    \midrule
    32  & 64  & 4 & 5.1 & 10.5 & --- \\
    64  & 64  & 2 & --- & ---  & 27.4 \\
    64  & 128 & 4 & 7.4 & 19.8 & 36.1 \\
    128 & 128 & 4 & 8.7 & ---  & 53.6 \\
    128 & 256 & 2 & --- & ---  & 96.0 \\
    256 & 256 & 4 & 9.5 & 36.2 & --- \\
    \bottomrule
  \end{tabular}
\end{table}

The gap between a workload's aggregate and peak speedup reflects its spread of
problem sizes: the peak comes from the largest, best-amortized case, while the
aggregate also folds in the smaller, overhead-dominated cases.

\subsection{Scaling Behavior}
The representative cases make two trends visible across the suite. First, wider meshes help more on larger
problems: moving from $2\times2$ to $8\times8$ multiplies the available MAC lanes
by $16\times$, but that capacity only translates into speedup when a tile is
large enough to keep the mesh busy, so the largest gains appear on the biggest
matmul, attention, and prefill cases, where the $8\times8$ curve pulls furthest
ahead. Second, small problems are overhead-dominated: fixed descriptor handling,
DMA setup, and controller scheduling are amortized poorly when there is little
compute per launch, which is why the smaller configurations improve over Rocket
but saturate earlier and why aggregate speedups (which include every small case)
sit below the peaks. Causal attention is slower than its non-causal counterpart:
masking removes useful work from part of the score matrix while the control flow
must still preserve the causal dependency structure. Overall, the results confirm that a single shared-mesh design scales
smoothly across an order of magnitude of lane count while covering both matrix
multiplication and full attention.

\section{Preliminary Area Study}
\label{sec:vlsi}

We use a preliminary Hammer/OpenROAD study on Sky130 to estimate standalone
\texttt{AtikCore} area. It excludes RocketTile, ChipTop, post-route timing,
power, and silicon validation; it should therefore be interpreted only as an
early area indication for the accelerator block.
Table~\ref{tab:vlsi-overview} reports the cell count, area, and area per MAC
lane for all three mesh sizes.

\begin{table}[H]
  \centering
  \caption{Preliminary standalone \texttt{AtikCore} area estimate
    ($\mu\mathrm{m}^2$) across mesh configurations.}
  \label{tab:vlsi-overview}
  \begin{tabular}{lrrr}
    \toprule
    & $2\times2$ & $4\times4$ & $8\times8$ \\
    \midrule
    MAC lanes            & 4       & 16        & 64 \\
    Cells                & 114{,}575 & 190{,}722 & 477{,}464 \\
    Estimated area       & 721{,}212 & 1{,}217{,}184 & 3{,}130{,}859 \\
    Area / MAC lane      & 180{,}303 & 76{,}074  & 48{,}920 \\
    \bottomrule
  \end{tabular}
\end{table}

In Table~\ref{tab:vlsi-overview}, the estimated area per MAC lane falls as fixed control and DMA infrastructure is
amortized over larger meshes. This observation is descriptive only: without
timing, power, and complete-chip results, it is not a PPA comparison.

Figure~\ref{fig:vlsi-layout} shows implementation views from the preliminary
standalone VLSI flow. These views complement the area estimates by illustrating
the physical organization of the accelerator.

\clearpage
\begin{strip}
\begin{minipage}{\textwidth}
  \centering
  \includegraphics[width=\linewidth]{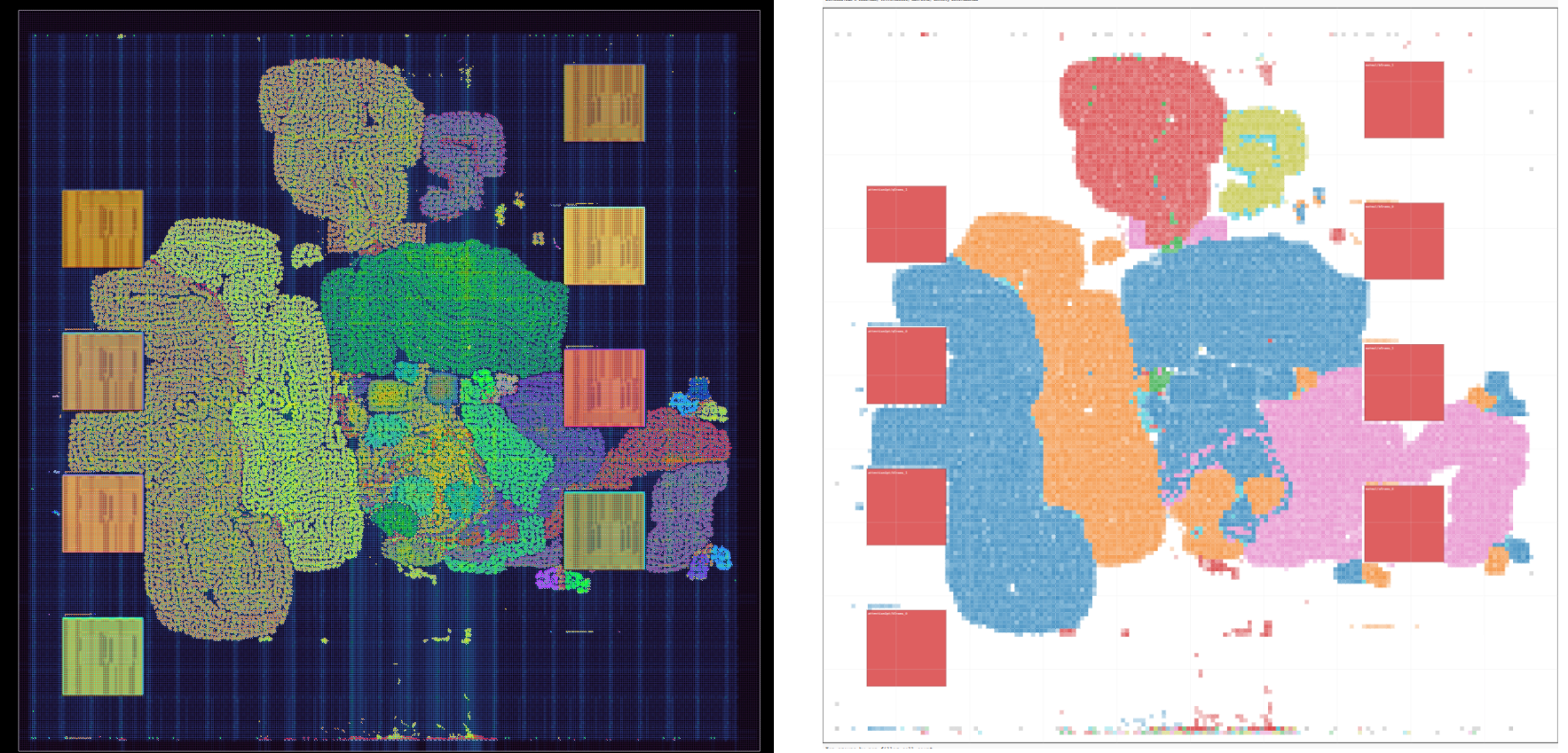}
  \captionof{figure}{Preliminary standalone AtikCore VLSI output from the
    Sky130 flow: layout view (left) and colored placement view (right).}
  \label{fig:vlsi-layout}
\end{minipage}
\end{strip}

\section{Discussion and Limitations}
\label{sec:discussion}

\paragraph{Baseline.}
The reported speedups are measured against the same in-order Rocket core running
BF16 software kernels. This deliberately isolates the effect of the accelerator
on a fixed RISC-V system, but Rocket is a scalar core without a SIMD or vector
unit, so part of the large absolute speedup reflects how poorly a scalar
pipeline handles dense BF16 matrix work and data-dependent softmax. The
comparison answers ``what does attaching Atik buy this core,'' not ``how does
Atik compare to a GPU or a vector machine''; those cross-platform comparisons are
future work.

\paragraph{Scalar-scheduled softmax.}
Atik keeps $QK^\top$ and $PV$ on the shared mesh but runs the online-softmax
bookkeeping and final normalization on a single scalar lane
(Section~\ref{sec:attention}). This is an area-first choice, but
it also serializes the non-linear part of attention. On very large score tiles
this scalar region becomes the dominant sequential cost. The number of scalar
lanes and LUT instances is a hardware parameter, not an ABI feature, so a
throughput-oriented instance can widen it without changing accelerator software;
quantifying that trade-off is left to future work.

\paragraph{Dense, non-sparse attention.}
Atik accelerates dense attention and does not exploit sparsity. Sparse-attention
accelerators skip low-contribution key/value rows and can achieve higher
effective throughput on workloads with exploitable sparsity, at the cost of
approximation and, often, retraining. Atik trades that opportunity for
generality and a BF16-compatible interface with measured benchmark error; adding an optional
sparsity path is complementary rather than contradictory.

\paragraph{Implementation scope.}
The preliminary area study targets standalone \texttt{AtikCore} on Sky130; it
does not include RocketTile, ChipTop, routing closure, or tapeout. Its area
estimate is not a silicon or PPA result. Extending the flow to a complete tile
and reporting timing and power are natural
next steps. Finally, the workload study covers matmul,
attention, and single-layer ViT, TinyBERT, and GPT-2 prefill; the autoregressive
decode stage with a KV cache is not yet accelerated and is an important target
for future work.

\section{Conclusion}
\label{sec:conclusion}

We presented Atik, an open-source, tightly-coupled transformer accelerator for
the RISC-V RocketChip. Atik attaches over RoCC, is driven by a minimal
descriptor-based software interface, and accelerates both of the recurring
transformer bottlenecks---dense matrix multiplication and attention---with BF16
input/output. Its central idea is a single shared fixed-point MAC mesh that serves
matmul, $QK^\top$ scoring, and probability--value accumulation, wrapped in a
streaming online-softmax schedule that never materializes the full score matrix.
Evaluated cycle-accurately on an AWS~F2 FPGA through FireSim, Atik reaches up to
$225.8\times$ speedup on attention and $31.5\times$ on matmul over the Rocket
baseline, with hybrid projected model speedups of $44\times$ on ViT and
$32\times$ on GPT-2 prefill. The preliminary area estimate shows the expected
amortization of shared control as the mesh grows, but it is not a PPA result.
Together these results show how a shared-mesh, BF16-interface,
tightly-coupled accelerator can reduce transformer-kernel cycles in the open
RISC-V ecosystem.

\paragraph{Availability.}
Atik, including its hardware, software, and physical-design flow, is available
at \href{https://github.com/AhmedZeer/atik}{\texttt{github.com/AhmedZeer/atik}}.

\bibliographystyle{ACM-Reference-Format}
\bibliography{references}

@INPROCEEDINGS{gemmini,
  author={Genc, Hasan and Kim, Seah and Amid, Alon and Haj-Ali, Ameer and Iyer, Vighnesh and Prakash, Pranav and Zhao, Jerry and Grubb, Daniel and Liew, Harrison and Mao, Howard and Ou, Albert and Schmidt, Colin and Steffl, Samuel and Wright, John and Stoica, Ion and Ragan-Kelley, Jonathan and Asanovic, Krste and Nikolic, Borivoje and Shao, Yakun Sophia},
  booktitle={2021 58th ACM/IEEE Design Automation Conference (DAC)}, 
  title={Gemmini: Enabling Systematic Deep-Learning Architecture Evaluation via Full-Stack Integration}, 
  year={2021},
  volume={},
  number={},
  pages={769-774},
  doi={10.1109/DAC18074.2021.9586216}
}

@ARTICLE{chipyard,
  author={Amid, Alon and Biancolin, David and Gonzalez, Abraham and Grubb, Daniel and Karandikar, Sagar and Liew, Harrison and Magyar, Albert and Mao, Howard and Ou, Albert and Pemberton, Nathan and Rigge, Paul and Schmidt, Colin and Wright, John and Zhao, Jerry and Shao, Yakun Sophia and Asanović, Krste and Nikolić, Borivoje},
  journal={IEEE Micro}, 
  title={Chipyard: Integrated Design, Simulation, and Implementation Framework for Custom SoCs}, 
  year={2020},
  volume={40},
  number={4},
  pages={10-21},
  doi={10.1109/MM.2020.2996616}}

@INPROCEEDINGS{firesim,
  author={Karandikar, Sagar and Mao, Howard and Kim, Donggyu and Biancolin, David and Amid, Alon and Lee, Dayeol and Pemberton, Nathan and Amaro, Emmanuel and Schmidt, Colin and Chopra, Aditya and Huang, Qijing and Kovacs, Kyle and Nikolic, Borivoje and Katz, Randy and Bachrach, Jonathan and Asanovic, Krste},
  booktitle={2018 ACM/IEEE 45th Annual International Symposium on Computer Architecture (ISCA)}, 
  title={FireSim: FPGA-Accelerated Cycle-Exact Scale-Out System Simulation in the Public Cloud}, 
  year={2018},
  volume={},
  number={},
  pages={29-42},
  doi={10.1109/ISCA.2018.00014}}

@INPROCEEDINGS{vexp,
  author={Wang, Run and Islamoglu, Gamze and Belano, Andrea and Potocnik, Viviane and Conti, Francesco and Garofalo, Angelo and Bonini, Luca},
  booktitle={2025 IEEE 32nd Symposium on Computer Arithmetic (ARITH)}, 
  title={VEXP: A Low-Cost RISC-V ISA Extension for Accelerated Softmax Computation in Transformers}, 
  year={2025},
  volume={},
  number={},
  pages={37-44},
  doi={10.1109/ARITH64983.2025.00016}
}

@INPROCEEDINGS{ita,
  author={Islamoglu, Gamze and Scherer, Moritz and Paulin, Gianna and Fischer, Tim and Jung, Victor J.B. and Garofalo, Angelo and Benini, Luca},
  booktitle={2023 IEEE/ACM International Symposium on Low Power Electronics and Design (ISLPED)}, 
  title={ITA: An Energy-Efficient Attention and Softmax Accelerator for Quantized Transformers}, 
  year={2023},
  volume={},
  number={},
  pages={1-6},
  doi={10.1109/ISLPED58423.2023.10244348}}

@misc{onlinesoftmax,
      title={Online normalizer calculation for softmax}, 
      author={Maxim Milakov and Natalia Gimelshein},
      year={2018},
      eprint={1805.02867},
      archivePrefix={arXiv},
      primaryClass={cs.PF},
      url={https://arxiv.org/abs/1805.02867}, 
}

@INPROCEEDINGS{softermax,
  author={Stevens, Jacob R. and Venkatesan, Rangharajan and Dai, Steve and Khailany, Brucek and Raghunathan, Anand},
  booktitle={2021 58th ACM/IEEE Design Automation Conference (DAC)}, 
  title={Softermax: Hardware/Software Co-Design of an Efficient Softmax for Transformers}, 
  year={2021},
  volume={},
  number={},
  pages={469-474},
  doi={10.1109/DAC18074.2021.9586134}}

@INPROCEEDINGS{googletpu,
  author={Jouppi, Norman P. and Young, Cliff and Patil, Nishant and Patterson, David and Agrawal, Gaurav and Bajwa, Raminder and Bates, Sarah and Bhatia, Suresh and Boden, Nan and Borchers, Al and Boyle, Rick and Cantin, Pierre-luc and Chao, Clifford and Clark, Chris and Coriell, Jeremy and Daley, Mike and Dau, Matt and Dean, Jeffrey and Gelb, Ben and Ghaemmaghami, Tara Vazir and Gottipati, Rajendra and Gulland, William and Hagmann, Robert and Ho, C. Richard and Hogberg, Doug and Hu, John and Hundt, Robert and Hurt, Dan and Ibarz, Julian and Jaffey, Aaron and Jaworski, Alek and Kaplan, Alexander and Khaitan, Harshit and Killebrew, Daniel and Koch, Andy and Kumar, Naveen and Lacy, Steve and Laudon, James and Law, James and Le, Diemthu and Leary, Chris and Liu, Zhuyuan and Lucke, Kyle and Lundin, Alan and MacKean, Gordon and Maggiore, Adriana and Mahony, Maire and Miller, Kieran and Nagarajan, Rahul and Narayanaswami, Ravi and Ni, Ray and Nix, Kathy and Norrie, Thomas and Omernick, Mark and Penukonda, Narayana and Phelps, Andy and Ross, Jonathan and Ross, Matt and Salek, Amir and Samadiani, Emad and Severn, Chris and Sizikov, Gregory and Snelham, Matthew and Souter, Jed and Steinberg, Dan and Swing, Andy and Tan, Mercedes and Thorson, Gregory and Tian, Bo and Toma, Horia and Tuttle, Erick and Vasudevan, Vijay and Walter, Richard and Wang, Walter and Wilcox, Eric and Yoon, Doe Hyun},
  booktitle={2017 ACM/IEEE 44th Annual International Symposium on Computer Architecture (ISCA)}, 
  title={In-datacenter performance analysis of a tensor processing unit}, 
  year={2017},
  volume={},
  number={},
  pages={1-12},
  doi={10.1145/3079856.3080246}}

@ARTICLE{sysArrKung,
  author={Kung},
  journal={Computer}, 
  title={Why systolic architectures?}, 
  year={1982},
  volume={15},
  number={1},
  pages={37-46},
  doi={10.1109/MC.1982.1653825}}

@techreport{rocketSoC,
    Author= {Asanović, Krste and Avizienis, Rimas and Bachrach, Jonathan and Beamer, Scott and Biancolin, David and Celio, Christopher and Cook, Henry and Dabbelt, Daniel and Hauser, John and Izraelevitz, Adam and Karandikar, Sagar and Keller, Ben and Kim, Donggyu and Koenig, John and Lee, Yunsup and Love, Eric and Maas, Martin and Magyar, Albert and Mao, Howard and Moreto, Miquel and Ou, Albert and Patterson, David A. and Richards, Brian and Schmidt, Colin and Twigg, Stephen and Vo, Huy and Waterman, Andrew},
    Title= {The Rocket Chip Generator},
    Year= {2016},
    Month= {Apr},
    Url= {http://www2.eecs.berkeley.edu/Pubs/TechRpts/2016/EECS-2016-17.html},
    Number= {UCB/EECS-2016-17},
}

@ARTICLE{energon,
  author={Zhou, Zhe and Liu, Junlin and Gu, Zhenyu and Sun, Guangyu},
  journal={IEEE Transactions on Computer-Aided Design of Integrated Circuits and Systems}, 
  title={Energon: Toward Efficient Acceleration of Transformers Using Dynamic Sparse Attention}, 
  year={2023},
  volume={42},
  number={1},
  pages={136-149},
  doi={10.1109/TCAD.2022.3170848}}

@INPROCEEDINGS{a3,
  author={Ham, Tae Jun and Jung, Sung Jun and Kim, Seonghak and Oh, Young H. and Park, Yeonhong and Song, Yoonho and Park, Jung-Hun and Lee, Sanghee and Park, Kyoung and Lee, Jae W. and Jeong, Deog-Kyoon},
  booktitle={2020 IEEE International Symposium on High Performance Computer Architecture (HPCA)}, 
  title={{A\textsuperscript{3}}: Accelerating Attention Mechanisms in Neural Networks with Approximation}, 
  year={2020},
  volume={},
  number={},
  pages={328-341},
  doi={10.1109/HPCA47549.2020.00035}}

@INPROCEEDINGS{ticsat,
  author={Amirshahi, Alireza and Harrison Klein, Joshua Alexander and Ansaloni, Giovanni and Atienza, David},
  booktitle={2023 28th Asia and South Pacific Design Automation Conference (ASP-DAC)}, 
  title={TiC-SAT: Tightly-coupled Systolic Accelerator for Transformers}, 
  year={2023},
  volume={},
  number={},
  pages={657-663},
  doi={}}

@inproceedings{sanger,
  title={Sanger: A Co-Design Framework for Enabling Sparse Attention using Reconfigurable Architecture},
  author={Lu, Liqiang and Xie, Siyu and Liao, Yang and others},
  booktitle={2021 54th Annual IEEE/ACM International Symposium on Microarchitecture (MICRO)},
  pages={777--788},
  year={2021},
  doi={10.1145/3466752.3480125},
  url={https://doi.org/10.1145/3466752.3480125}
}

@inproceedings{elsa,
  title={ELSA: Hardware-Software Co-design for Efficient, Lightweight Self-Attention Mechanism in Neural Networks},
  author={Ham, Tae Jun and Lee, Yejin and Seo, Seong Hoon and Kim, Soosung and Choi, Hyunji and Jung, Sung Jun and Lee, Jae W.},
  booktitle={2021 ACM/IEEE 48th Annual International Symposium on Computer Architecture (ISCA)},
  year={2021},
  doi={10.1109/ISCA52012.2021.00060},
  url={https://doi.org/10.1109/ISCA52012.2021.00060}
}

@inproceedings{swiftron,
  title={SwiftTron: An Efficient Hardware Accelerator for Quantized Transformers},
  author={Marchisio, Alberto and Dura, Davide and Capra, Maurizio and Martina, Maurizio and Masera, Guido and Shafique, Muhammad},
  booktitle={2023 International Joint Conference on Neural Networks (IJCNN)},
  year={2023},
  doi={10.1109/IJCNN54540.2023.10191521},
  url={https://doi.org/10.1109/IJCNN54540.2023.10191521}
}

@inproceedings{sole,
  title={SOLE: Hardware-Software Co-design of Softmax and LayerNorm for Efficient Transformer Inference},
  author={Wang, Wenxun and Zhou, Shuchang and Sun, Wenyu and Sun, Peiqin and Liu, Yongpan},
  booktitle={2023 IEEE/ACM International Conference on Computer-Aided Design (ICCAD)},
  year={2023},
  doi={10.1109/ICCAD57390.2023.10323725},
  url={https://doi.org/10.1109/ICCAD57390.2023.10323725}
}

@INPROCEEDINGS{chisellang,
  author={Bachrach, Jonathan and Vo, Huy and Richards, Brian and Lee, Yunsup and Waterman, Andrew and Avižienis, Rimas and Wawrzynek, John and Asanović, Krste},
  booktitle={DAC Design Automation Conference 2012}, 
  title={Chisel: Constructing hardware in a Scala embedded language}, 
  year={2012},
  volume={},
  number={},
  pages={1212-1221},
  doi={10.1145/2228360.2228584}}

@inproceedings{hammer,
author = {Liew, Harrison and Grubb, Daniel and Wright, John and Schmidt, Colin and Krzysztofowicz, Nayiri and Izraelevitz, Adam and Wang, Edward and Asanovi\'{c}, Krste and Bachrach, Jonathan and Nikoli\'{c}, Borivoje},
title = {Hammer: a modular and reusable physical design flow tool: invited},
year = {2022},
isbn = {9781450391429},
publisher = {Association for Computing Machinery},
address = {New York, NY, USA},
url = {https://doi.org/10.1145/3489517.3530672},
doi = {10.1145/3489517.3530672},
booktitle = {Proceedings of the 59th ACM/IEEE Design Automation Conference},
pages = {1335–1338},
numpages = {4},
location = {San Francisco, California},
series = {DAC '22}
}

\end{document}